# Geometry-Informed Material Recognition

Joseph DeGol        Mani Golparvar-Fard        Derek Hoiem
University of Illinois Urbana-Champaign
{degol2, mgolpar, dhoiem}@illinois.edu

## Abstract

*Our goal is to recognize material categories using images and geometry information. In many applications, such as construction management, coarse geometry information is available. We investigate how 3D geometry (surface normals, camera intrinsic and extrinsic parameters) can be used with 2D features (texture and color) to improve material classification. We introduce a new dataset, GeoMat, which is the first to provide both image and geometry data in the form of: (i) training and testing patches that were extracted at different scales and perspectives from real world examples of each material category, and (ii) a large scale construction site scene that includes 160 images and over 800,000 hand labeled 3D points. Our results show that using 2D and 3D features both jointly and independently to model materials improves classification accuracy across multiple scales and viewing directions for both material patches and images of a large scale construction site scene.*

## 1. Introduction

Our goal is to recognize material categories using images and estimated 3D points. In prior material recognition research, surface geometry is a confounder, and much effort goes into creating features that are stable under varying perspective (e.g., scale and rotationally invariant features [33]) and lighting. Although the resulting systems often perform well for standard material/texture datasets [11, 16, 5, 25], their success does not always translate to improved categorization in natural objects or scenes [22, 12]. However, for many important applications, 3D surface geometry can be estimated, rather than marginalized, and used to improve performance. For example, a ground robot can estimate surface geometry from stereo when identifying navigable terrain. Likewise, when surveying progress in a construction site, 3D points from LiDAR or structure-from-motion can help distinguish between concrete and stone to determine if a facade is in place. In principal, geometric estimates should help with material classification by revealing surface orientation and roughness and disambiguating texture cues, but because surface texture and geometry interact in com-plex ways, it is not clear how best to take advantage of 3D points. Can local geometry cues be simply added to existing color/texture features, or do they need to be considered jointly? Are approaches to improve robustness of texture descriptors still helpful? Is it helpful to rectify the image based on surface geometry? Our paper aims to answer these questions and provide a material recognition approach that is well-suited to applications for which surface geometry estimates are available.

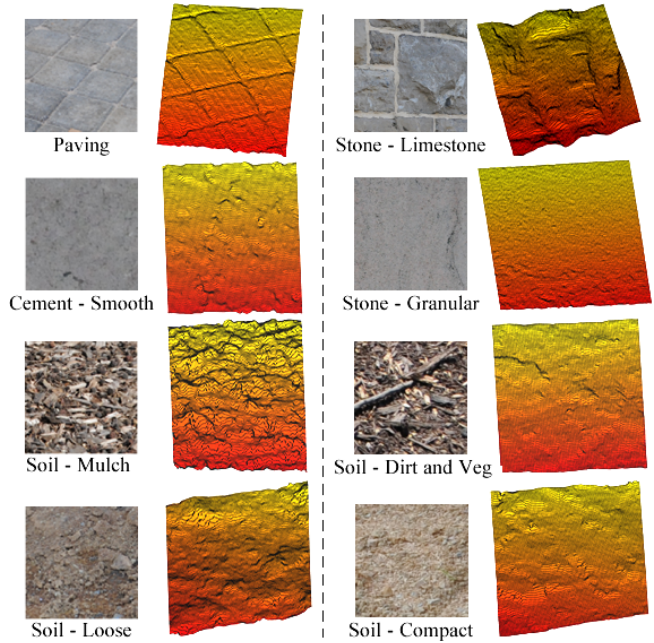

Figure 1: The material patches shown in column one were misclassified as the class shown in column three by [8] because the classes are visually similar. However, the geometry (column two and four) for these patches is different. This paper investigates how to use differences in 3D geometry to improve material classification. We also contribute the GeoMat dataset consisting of images and geometry for material patches and a large scale construction site scene.

We introduce a new dataset of construction materials photographed in natural outdoor lighting (called "GeoMat"

for geometry/materials). Many of the 19 material categories (Fig. 3) are highly confusable, such as "paving" vs. "limestone" or "smooth cement" vs. "granular stone" (Fig. 1), but these distinctions are important in a construction setting. For each category, several different physical samples are photographed from a variety of orientations and positions, and structure-from-motion [31] and multi-view stereo [14] are used to estimate 3D points. We explore two test settings: individual 2D/3D patches of material samples and scene-scale images of construction sites with 3D point clouds.

Using our GeoMat dataset, we investigate how estimated 3D geometry can improve material classification in real world scenes. Surface orientation and roughness provide valuable cues to material category, and we model them with histograms of surface normals. Additionally, observed texture is due to a combination of surface markings, micro-geometric texture, and camera-relative surface normal. Our geometric detail is not sufficient to model micro-geometric texture, but by jointly representing camera-relative surface normal and texture response, we may reduce ambiguity of signal. Thus, we try jointly representing texture and normals. An alternative strategy is to frontally warp the image, based on surface normal, which would undo perspective effects at the cost of some resolution due to interpolation. Our main technical contribution is to investigate all of these strategies to determine which strategy or combination of strategies makes the best use of geometric information. We also investigate how performance of 3D-sensitive features varies with scale and surface orientation.

In summary, our **contributions** are: (1) we create the GeoMat dataset for studying material categorization from images supplemented with sparse 3D points; (2) we investigate several strategies for using 3D geometry with color and texture to improve material recognition; (3) we investigate effects of scale and orientation and application to images of construction sites.

## 2. Related Work

**Features:** Early methods for material classification used filter bank responses to extract salient statistical characteristics from image patches [11, 37, 32, 4, 3]. Leung and Malik [21] introduced the LM-filter bank and proposed "3D Textons", which, despite the name, are clustered 2D filter responses at each pixel without direct 3D information. The term "texton" was coined by Julez [18] twenty years earlier to describe elements of human texture perception, and "3D" conveys the goal of classifying 3D material textures. Varma and Zisserman [33] later proposed the "RFS" filter bank and an in-plane rotationally invariant (via max pooling) "MR8" response set. A string of subsequent work, led by Varma and Zisserman, replaced filter responses with more direct clusterings and statistics of intensities of small pixel neighborhoods [34, 26, 36, 7, 15, 29, 30, 24]. Liu et al. [22] explored a variety of color, texture, gradient, and curvature features for classifying object-level material images. It was recently shown by Cimpoi et al. [8, 9] that convolutional neural networks and fisher vectors with dense SIFT outperforms previous approaches for texture classification.

These works all explore purely 2D image-based features and, as such, aim to be robust to 3D surface variations by encoding texture for samples observed from various viewpoints and lighting. We show that directly encoding local surface geometry both jointly and independently with texture yields significant gains. We are the first, to our knowledge, to investigate how to integrate 3D geometric cues with texture representations for material classification. We note that object segmentation from RGB-D images is a commonly studied problem (e.g., Koppula et al. [19]), but because the image resolution is too low for texture to be an effective cue and the focus is on object rather than material categories, the problem is dissimilar (the same is also true of LiDAR classification approaches).

**Datasets:** The CUReT dataset created by Dana et al. [11] was the first large-scale texture/material dataset, providing 61 material categories, photographed in 205 viewing and lighting conditions. The KTH-TIPS dataset by Hayman et al. [16] added scale variation by imaging 10 categories from the CUReT dataset at different scales. For both datasets, all images for a category were from the same physical sample, so that they may be more accurately called texture categorization than material categorization datasets. Subsequently, KTH-TIPS2 by Caputo et al. [5] was introduced, adding images from four physical samples per category. Still, variation of material complexity within categories was limited, motivating Liu et al. [22] to create the Flickr Materials Database containing images for ten categories with 50 material swatch images and 50 object-level images. Several recent datasets have focused on material recognition for applications. This includes the construction materials dataset by Dimitrov and Golparvar-Fard [12] which consists of 200x200 patches of 20 common construction materials, and the Describable Texture Dataset by Cimpoi et al. [8] which provides 5,640 texture images jointly annotated with 47 material attributes. Most recently, Bell et al. [2] contributed the Materials in Context Database, consisting of many full scenes with material labels.

While these datasets provide ample resources for studying image-based material classification, there does not yet exist a dataset that provides geometry information together with real-world material images. Our GeoMat dataset provides real world material images and geometric information in the form of point clouds, surface normals, and camera intrinsic and extrinsic parameters. Our dataset also differs in that the taxonomy is chosen to be relevant to a practical application (construction management), rather than based on visual distinctiveness, leading to several groups of highly confusable material types. For example, Varma and Zisserman [35] report accuracy of 96.4% on the 61 CUReT classes using the MR8 representation; the same MR8 representation achieves only 32.5% accuracy on our dataset.

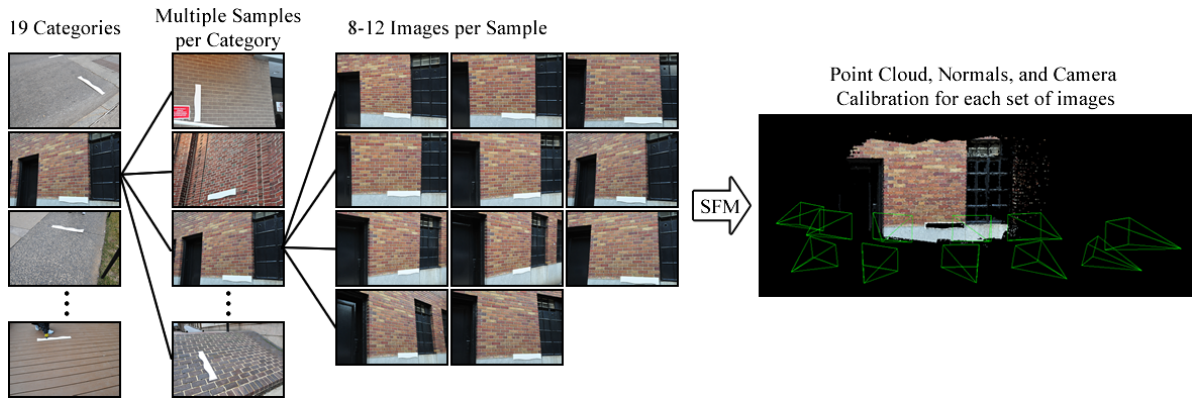

Figure 2: Each GeoMat material category is made from 3 to 26 different samples where each sample consists of 8 to 12 images at different viewpoints, a segmented point cloud, and normal vectors.

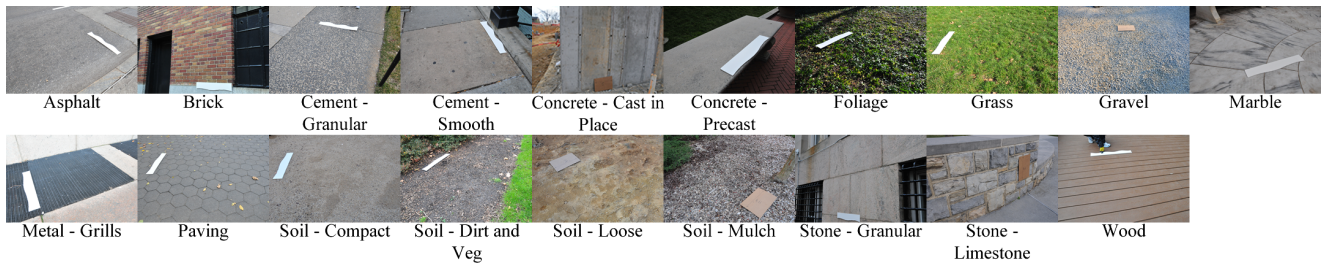

Figure 3: GeoMat represents 19 material categories.

## 3. Dataset

We created the GeoMat dataset (Figs. 2, 3, and 4) to investigate how local geometric data can be used with image data to recognize materials in real-world environments. The training set consists of "focus scale" 100x100 patches of single materials sampled from high resolution photographs of buildings and grounds. There are two test sets: (i) 100x100 patches sampled from photographs of different physical surfaces, and (ii) "scene scale" photographs of a construction site. Both focus scale and scene scale datasets consist of images and associated 3D points estimated through multiview 3D reconstruction.

### 3.1. Focus Scale Training and Testing Sets

The focus scale data is sampled from high-resolution (4288x2848 pixels) images that predominantly depict a single material, such as a "brick" wall or "soil - compact" ground. The dataset consists of 19 material categories as shown in Fig. 3. There are between 3 and 26 different physical surfaces (i.e. different walls or ground areas) for each category; each surface is photographed from 8 to 12 viewpoints (Fig. 2). A marker of known scale is present in each image. Structure from motion [31] and multi-view stereo [14] are used to generate a point cloud, normal vectors, and camera intrinsic and extrinsic parameters. The points are manually labeled into regions of interest to facilitate sampling patches that consist purely of one material.

We make training and testing splits by assigning approximately 70% of the physical surfaces of each category to training and the remainder to testing. For example, given a category with three surfaces, training samples will come from two of the surfaces and testing samples will come from the remaining unused surface. Similarly, for a category with 23 samples, training samples will come from 16 of the surfaces and testing samples will come from the remaining 7 unused surfaces. Since each category consists of at least three different surfaces, this ensures that there are at least two surfaces per category for training, at least one surface per category for testing, and the samples drawn for training are from different surfaces than those drawn for testing.

For each category, we extract 100 training patches and 50 testing patches at 100x100, 200x200, 400x400, and 800x800 resolutions. This results in a total of 400 training patches and 200 testing patches per category. These patches are scaled to 100x100 to simulate viewing the materials at different scales/distances. We extract an equal number of patches from each surface. For example, if we want to extract 200 testing patches from 10 surfaces, then 20 testing patches are extracted from each surface. Since each surface consists of many images, we then divide the intended number of patches evenly among the images of that surface. Continuing with the example, if a surface has 10 images and we want to extract 20 total patches from that surface, then we extract 2 patches per image. Each patch is then extracted randomly from within a region of the image that was manually annotated as representative of the intended category.

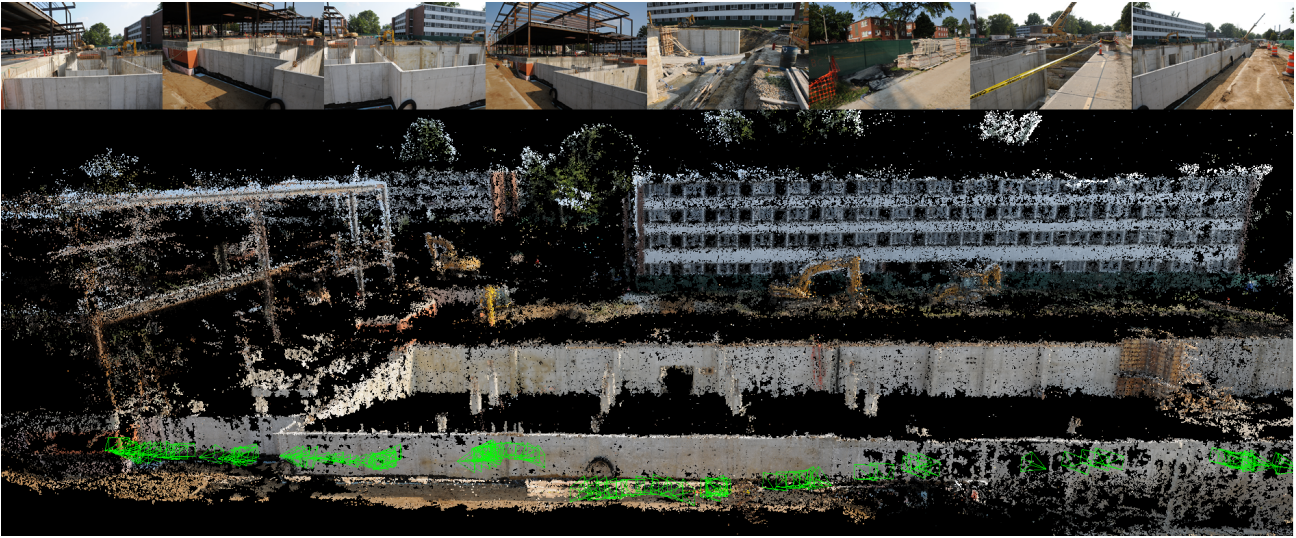

Figure 4: The scene scale dataset consists of 160 images of a construction site with an accompanying point cloud, normal vectors, and camera intrinsic and extrinsic parameters. The SFM-registered camera frusta are shown in green. 11 of the 19 material categories are represented: "Brick", "Cement – Smooth", "Concrete – Precast", "Concrete – Cast in Place", "Foliage", "Grass", "Gravel", "Metal – Grills", "Soil – Compact", "Soil – Loose", and "Wood".

Each patch consists of image data, geometry data, and from which category and surface it was drawn. Examples are shown in Fig. 1. Image data includes normalized grayscale and HSV images and the location in the image from which the sample was drawn. Geometry data includes a sparse depth map, sparse normal map, intrinsic and extrinsic camera parameters, gravity vector, and scale.

### 3.2. Scene Scale Testing Set

The scene scale data consists of 160 images (4288x2848 pixels each) of one large construction site. Of the 19 material categories, 11 are represented: "Brick", "Cement – Smooth", "Concrete – Precast", "Concrete – Cast in Place", "Foliage", "Grass", "Gravel", "Metal – Grills", "Soil – Compact", "Soil – Loose", and "Wood". Structure from motion and multi-view stereo were used to generate a point cloud, normal vectors, and camera intrinsic and extrinsic parameters. The point cloud is hand-labeled to match our 19 material categories. Points not matching one of the 19 categories are labeled as unknown. Fig. 4 provides a depiction of the scene scale testing set.

The scene scale data is used only for testing. We use the dataset to verify that our conclusions drawn from the simpler focus scale dataset still hold when classifying regions in more typical images. Others could use the data for testing multiview material recognition or transfering patch-based material models to scene-scale images. Labeled 3D points (826,509 total) that are viewable in a given image are back-projected onto pixels, so that a sparse set of pixels (about 21,500 per image on average) has ground truth labels in each image. When testing with the scene scale data, we use the entire focus scale dataset for training.

## 4. Classification with Geometric Features

### 4.1. Features and Modeling

Our main interest is in how to use patch geometry to improve or augment image features. We describe the 2D texture and color features that we use and then describe several cues that leverage the estimated depth and surface normals.

#### 4.1.1 2D Features

**RFS/MR8:** The intensity pattern of a material is a good cue for recognition [33, 21, 28, 10] because it encodes surface albedo patterns and small-scale shape. Consider brick: we expect to see grainy rectangular blocks separated by layers of mortar. Filter banks have proven useful for capturing these and other intensity patterns for material recognition. We use the RFS filter bank and derived MR8 responses described by Varma and Zisserman [35], which are shown to be effective on the CUReT dataset [33]. The RFS filter set contains first and second derivative filters at 6 orientations and 3 scales (36 filters) and Gaussian and Laplacian of Gaussian (LoG) filters at scale $\sigma = 10$ (2 filters). The MR8 filters are created by keeping only the maximum filter response across each set of orientations for a given scale, along with the two Gaussian/LoG filters. The MR8 filters are intended to provide robustness to surface orientation. In training, filter responses at each pixel are clustered into 10 clusters per category using k-means, following the standard texton approach [21]. The RFS and MR8 features are histograms of these textons (clustered filter responses), normalized to sum to one.

**FV:** SIFT [23] features offer an alternative method of capturing texture patterns and were used by Lui et.al. [22] for material recognition on the Flicker Materials Dataset. We quantize multi-scale dense SIFT features using the Improved Fisher Vector framework [27] as described by Cimpoi et. al. [8]. In training, the dimensionality of the dense SIFT features is reduced to 80 using PCA. The reduced dense SIFT features are then clustered into 256 modes using a Gaussian Mixture Model. The feature vectors are mean and covariance deviations from the GMM modes. The feature vectors are $\ell^2$ normalized and sign square-rooted as is standard for Improved Fisher Vectors.

**HSV:** Materials can be recognized by their color — grass is often green, bricks are often red and brown, and asphalt is often gray. We incorporate color by converting image patches to the HSV color space. The HSV pixels are then clustered into five clusters per category using k-means, and the resulting histograms are used as features.

**CNN:** Convolutional Neural Networks offer another approach for capturing texture and color patterns. We follow the approach of Cimpoi et. al. [8, 9], and use the pre-trained VGG-M network of [20]. The features are extracted from the last convolutional layer of the network.

### 4.1.2 3D Features

We investigate three strategies for including 3D geometric information for material classification: (i) jointly cluster texture features and 3D normal vectors at each pixel (-N); (ii) independently cluster normal vectors, build histograms ($N_{3D}$), and add them to 2D features; and (iii) frontally rectify the image based on a plane fit before computing texture filter responses.

**-N:** Image texture is affected by albedo patterns, surface orientation, and small surface shape variations. These factors make classification based on filter responses more difficult. A common solution is to make features robust to surface orientation by learning from many examples or creating rotationally invariant features (as in MR8 and SIFT). We hypothesize that explicitly encoding geometry jointly with the texture features will be more discriminative.

We interpolate over the sparse 3D normal map to produce a pixel-wise estimate of normals for a given image patch. We then transform the normal vectors according to the camera calibration information so that the normals are in the coordinate frame of the image plane. For MR8 and RFS, we then concatenate the normal vectors onto the filter responses at each pixel and cluster them into 10 clusters per category to create MR8-N and RFS-N textons. The textons are then used to build MR8-N and RFS-N histograms. For FV, we first reduce the dimensionality of the SIFT features to 80 using PCA. Then, we concatenate the 3D normal vectors onto the reduced SIFT descriptors for each pixel and cluster into 256 modes using a Gaussian Mixture Model. The modes include characteristics of both the texture and normal vectors. The Improved Fisher Vector formulation [27] is then used to create FV-N feature vectors.

**$N_{3D}$:** It is unclear whether a joint or independent representation of geometry will perform better, and it is also possible that both representations may help with overall discrimination. Thus, we formulate the $N_{3D}$ feature as an independent representation of the sparse normal map.

As described for (-N), we interpolate over the sparse 3D normal map to produce pixel-wise normal estimates for each patch and transform the normal vectors into the coordinate frame of the image plane. Rather than concatenating the normals with the texture features (as was done with (-N)), we independently cluster the normal vectors into five clusters per category using k-means and use the resulting histograms as our $N_{3D}$ features. Note that we also tried clustering the normal vectors using a Gaussian Mixture Model and building Fisher Vectors but saw worse performance using this method.

**Rectification:** In addition to directly encoding 3D surface geometry, frontally rectifying the image may improve texture features by making filter responses more directly correspond to albedo and microshape changes, removing the confounding factor of overall surface orientation and scale. We perform rectification using a homography defined by making the mean surface normal face the camera. The rectified patch is scaled to 100x100.

### 4.2. Classification

For this work, we are interested in investigating the utility of different geometric features and establishing a base-line for classification with the GeoMat dataset. We use a one vs. all SVM scheme for classification because SVMs have been shown to achieve exemplary performance on texture classification tasks for all our 2D features [16, 8, 9]. Experiments with only histogram features (RFS/MR8, HSV, RFS-N/MR8-N, N$_{3D}$) benefit from weighting the histograms before concatenating them. We learn the weights by grid search using leave-one-out cross-validation on the training set with a nearest neighbor classifier (which can be done very efficiently by caching inter-example histogram distances for each feature type). The weighted and concatenated histograms are then classified with a $\chi^2$ SVM. For experiments that include non-histogram feature vectors (FV, FV-N, CNN), the feature vectors and histograms are individually L2 normalized before being concatenated. We use libSVM [6] for training.

| Features | - | +HSV | +N$_{3D}$ | +HSV+N$_{3D}$ |
|---|---|---|---|---|
| ( RFS [33] / RFS-N ) | ( 33.24 / 37.76 ) | ( 45.03 / 47.89 ) | ( 49.68 / 49.55 ) | ( 51.24 / 52.29 ) |
| ( MR8 [33] / MR8-N ) | ( 32.47 / 41.34 ) | ( 45.32 / 47.84 ) | ( 49.74 / 50.63 ) | ( 53.03 / 53.37 ) |
| ( FV [8] / FV-N ) | ( 60.97 / 66.95 ) | ( 62.92 / 68.76 ) | ( 65.87 / 68.16 ) | ( 66.37 / 69.05 ) |
| ( FV+CNN [8] / FV-N+CNN ) | ( 68.92 / 73.80 ) | ( 67.82 / 72.05 ) | ( 72.08 / **73.84** ) | ( 70.79 / 72.13 ) |

Table 1: Including 3D geometry features increases the mean accuracy for all feature sets. Both joint and independent modeling of the 3D geometry improve the mean accuracy. The best mean accuracy is 73.84%.

### 4.3. Application to Scene Scale

In our scene scale test set, the input is RGB images at original scale with a sparse set of reconstructed points. We use this data to verify that our conclusions on the curated focus scale dataset still hold for typical images of large-scale scenes. To apply our patch-based classifer, we segment each image into 290-300 superpixels (roughly 200x200 pixels each) using SLIC [1]. For each superpixel, we extract the image patch and corresponding sparse normal map for the minimum bounding rectangle. The sparse normal map is then interpolated and transformed into the coordinate frame of the image plane. The image patches are resized for the CNN and used as-is for all other features. Classification is done on each patch independently and accuracy is measured as the average accuracy per pixel label.

## 5. Results and Analysis

Table 1 provides the mean classification accuracies on the testing data of the focus scale component of the GeoMat dataset. Since jointly clustering texture and 3D geometry (-N) is an alternative representation of the texture features, we display it in conjunction with the texture representation (texture representation / joint texture and normal representation). Then, each extra feature set that is concatenated is shown as another column of the table. We consider all of the original 2D features (RFS, MR8, FV, FV+CNN) to be baselines. From this table we see that the highest overall accuracy is 73.84% for FV-N+CNN+N$_{3D}$ which outperforms the best 2D baseline of FV+CNN [8] at 68.92%. Note also that the accuracy of using just N$_{3D}$ features is 32.50%.

We also tried several other baselines. First, we tried the approach of Cimpoi et al. [9]. This approach constructs Improved Fisher Vectors from the output features of the last convolutional layer of the pre-trained ImageNet network [20]. This method achieved 63.79% mean accuracy on our focus scale dataset. We also investigated the texture classification method provided by Sifre et al. [30]. This method learns a joint rotation and translation invariant representation of image patches using a cascade of wavelet modulus operators implemented in a deep convolutional network. We tested this baseline method with the same range of octave options as [30] and achieved a best accuracy of 36.53% with the number of octaves set to 3. Both of these baselines performed worse than FV+CNN at 68.92%.

The results shown here are a subset of our experiments and were chosen to highlight interesting trends. Additional experiments can be found in the supplemental material.

**Both joint and independent representations of geometry improve mean classification accuracy.** These two options map to (-N) features and N$_{3D}$ features respectively. From Table 1 and comparing column by column, we first see in column two that the (-N) significantly improves the mean classification accuracy compared to the 2D texture features (e.g. FV-N outperforms FV). In column three, we see that HSV provides a boost to the mean accuracies in every case; however, the inclusion of (-N) still improves the mean accuracies by at least 2% and by almost 6% for FV+HSV (e.g. FV-N+HSV outperforms FV+HSV). In column four, we can make two observations. First, we see that the inclusion of independent normal features (N$_{3D}$) significantly improves the mean accuracy compared to the 2D texture features (e.g. FV+N$_{3D}$ outperforms FV). In addition, we see that in every case except RFS, including both joint and independent geometry features (-N and N$_{3D}$) improves over using just one (e.g. FV-N+N$_{3D}$ outperforms FV-N and FV+N$_{3D}$). Note that the improvement for adding either (-N) or N$_{3D}$ (e.g. FV-N or FV+N$_{3D}$) is larger than the additional improvement gained by adding one to the

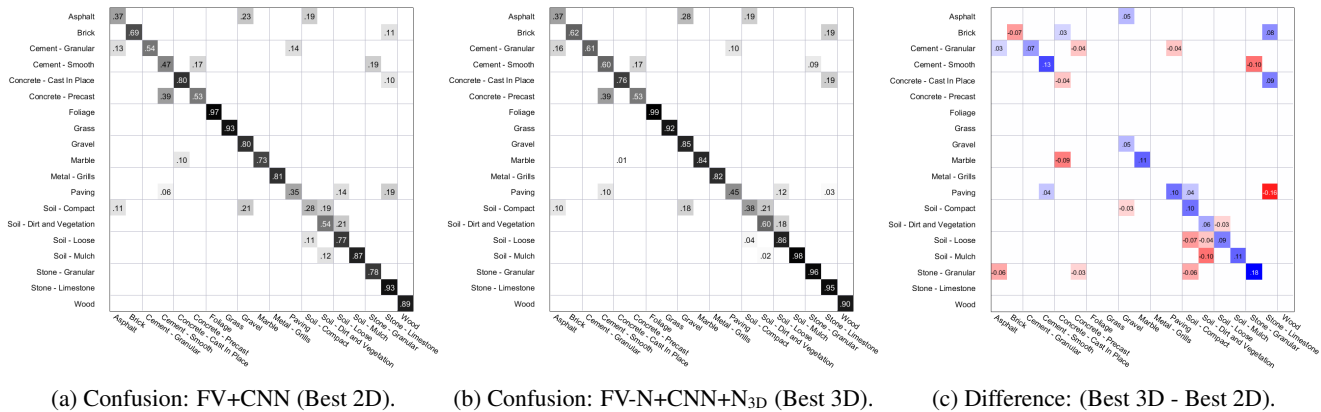

(a) Confusion: FV+CNN (Best 2D).  (b) Confusion: FV-N+CNN+N$_{3D}$ (Best 3D).  (c) Difference: (Best 3D - Best 2D).

Figure 5: The difference confusion matrix (Best 3D - Best 2D) shows the categories where the best 3D confusion matrix (b) performed better (blue cells) or worse (red cells) than the best 2D confusion matrix (a). The largest improvements are for Soil, Stone, and Cement. These categories often have similar visual appearance, but not necessarily similar 3D geometry. Including 3D geometry alleviates some of the confusion between these categories.

| Features | - | Rectified |
|---|---|---|
| RFS | 33.24 | 34.42 |
| RFS-N | 37.76 | 39.82 |
| MR8 | 32.47 | 35.03 |
| MR8-N | 41.34 | 42.05 |
| FV | 60.97 | 60.26 |
| FV-N | 66.95 | 66.82 |
| **FV+CNN** | 68.92 | 70.13 |
| **FV-N+CNN** | 73.80 | 72.97 |
| FV+**CNN** | 68.92 | 68.95 |
| FV-N+**CNN** | 73.80 | 73.71 |
| **FV+CNN** | 68.92 | 70.13 |
| **FV-N+CNN** | 73.80 | 72.92 |

Table 2: Rectification tends to help for filter features and not help when (-N) is included. Because the better performing features often perform worse with rectification, rectification does not appear to be an effective use of 3D geometry for improving classification. For FV+CNN, we denote which features are using rectification using boldfaced text.

other (e.g. adding N$_{3D}$ to FV-N). This makes sense because both features are modeling similar information; however, it is interesting that they still both contribute when used together. This trend is maintained with the inclusion of HSV features in column five.

**It is not clearly helpful to rectify the images based on surface geometry.** Table 2 shows the mean accuracies of the data with and without rectification. It is possible to apply the rectification to either FV or CNN; thus, we denote which features are using rectification using boldfaced text. From the results, we can see that rectification tends to improve the filter features (RFS, RFS-N, MR8, MR8-N) and some cases where (-N) is not included (FV+CNN). Rectification worsens the results for FV and also for most cases where (-N) is included (FV-N and FV-N+CNN). Because improvements are minimal when they exist and better performing feature combinations are often worse with rectification, we conclude that rectification is not an effective use of 3D geometry for improving classification. Since experiments with HSV and N$_{3D}$ do not use rectified images, we do not include them in this table, but the same trend continues and can be seen in the supplementary material.

**3D geometry helps with categories that look the same visually but have different 3D geometry.** Fig. 5a and Fig. 5b are the confusion matrices of the best performing 2D (FV+CNN) and 3D (FV-N+CNN+N$_{3D}$) feature sets respectively. For clarity, cells are hidden if they have a value below 0.1 in both confusion matrices. Fig. 5c is the subtraction of the best 2D confusion matrix from the best 3D confusion matrix. For clarity, cells are hidden if they have a value below 0.02.

The difference confusion matrix in Fig. 5c shows the categories where the best 3D confusion matrix performed better (blue cells) or worse (red cells) than the best 2D confusion matrix. The values along the diagonal (which represent improved classification accuracy for a given category) have improved in most cases. The largest improvements are for Soil (Compact, Dirt and Veg, Loose, and Mulch), Stone (Granular and Limestone), and Cement (Granular and Smooth). The reason we see larger gains in this area is because these materials look similar in terms of color and texture, but not similar in terms of their normal maps. In Fig. 1, we show in the two left-most columns of each row an example (image patch and normal map) that was misclassified in 2D but was correctly classified in 3D. The 2D incorrect guess then defines the class for columns three and four, and a hand-selected example is chosen from the training data that illustrates the possible similarity between image patches of the confused classes. It is clear from the examples shown in Fig. 1 why confusions are likely and also how the 3D geometry helps to alleviate these confusions. In particular, we see the flat panels and grooves for paving, the large stone outlines and mortar for limestone, the smooth surface of granular stone, and varying degrees of relief for the different types of soil (mulch, dirt and veg, loose, and compact).

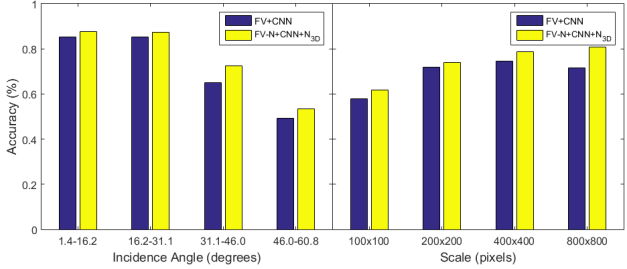

Figure 6: Mean accuracy improves as the incidence angle approaches a frontal view. Mean accuracy improves as scale increases. FV-N+CNN+$N_{3D}$ (yellow, Best 3D) outperforms FV+CNN (purple, Best 2D) for all scales and angles.

| Features | Pixel Labeling Accuracy |
|---|---|
| Hoiem et al. [17] | 18.53 |
| FV+CNN [8] | 21.01 |
| FV-N+CNN+$N_{3D}$ | 35.87 |

Table 3: For our scene scale dataset, the best 3D geometry feature set (FV-N+CNN+$N_{3D}$) outperforms the best 2D feature set (FV+CNN) and the external baseline, which is consistent with our results on the focus scale dataset.

**Including 3D geometry improves classification accuracy for all scales and viewing directions.** Fig. 6 shows the accuracy of the mean material classification as it depends on incidence angle and scale. It is interesting to see that there is a general improvement in accuracy for increased scale. We suspect this is because the texture pattern of certain material categories becomes more evident for farther scales (e.g. it is easier to see the layers of brick and mortar). We also see that the smaller incidence angles (closer to being a frontal view) have higher mean classification accuracies; however, the decrease in mean classification accuracy does not occur until we reach angles larger than 31.1 degrees. Lastly, it is worth noting that the best 3D features (FV-N+CNN+$N_{3D}$) improve over the best 2D features (FV+CNN) for all angles and scales.

**Results are consistent for the scene scale data.** Finally, we test on the scene scale component of the GeoMat dataset. Results are shown in Table 3. We chose to test the approach using the best performing 2D (FV+CNN) and 3D (FV-N+CNN+$N_{3D}$) feature sets from Table 1. The 3D feature set outperforms the 2D feature set considerably (35.87% vs. 21.01%), which is consistent with our results for the focus scale component of the GeoMat dataset.

As an external baseline, we train the superpixel-based classifier from Hoiem et al. [17] that includes region shape, color, and texture cues. The classifier is trained on our focus scale training set and applied to Felzenszwalb and Huttenlocher [13] superpixels generated from the test images, as in their original algorithm. The baseline classifier achieves 18.53% accuracy, which is slightly worse than our 2D features and much worse than our 3D features. Note that our approach and the baseline do not benefit from scene context or image position, which can be valuable cues, because they are trained using focus scale patches. Other constraints and priors could be used to obtain the best possible performance, but our experiments are intended to focus on the impact of geometric features on appearance models.

## 6. Conclusion

In this paper, we investigate how 3D geometry can be used to improve material classification. We include 3D geometry features with state-of-the-art material classification 2D features and find that both jointly and independently modeling 3D geometry improves mean classification accuracy. We also find that frontal rectification based on average surface normal is not an effective use of 3D geometry for material classification. We also contribute the GeoMat dataset which consists of image and geometry data for isolated walls and ground areas and a large scale construction site scene. Directions for future work include taking advantage of multi view and contextual constraints when inferring material for large scale scenes from photo collections.

### ACKNOWLEDGMENT

This work is supported by NSF Grant CMMI-1446765 and the DoD National Defense Science and Engineering Graduate Fellowship (NDSEG). We gratefully acknowledge the support of NVIDIA Corporation with the donation of the Tesla K40 GPUs used for this research. Thanks to Andrey Dimitrov, Banu Muthukumar, and Simin Liu for help with data collection.